\documentclass{article}
\usepackage{a4}
\usepackage{colacl}
\usepackage[T1]{tipa}
\usepackage{graphicx}

\title{\raisebox{0.75\baselineskip}[0in][0in]{\parbox{\linewidth}{\small\tt\begin{flushright}
        {\bf In:} Eisner, J., L. Karttunen and A.
        Th\'{e}riault (eds.), {\em Finite-State Phonology:\@
          Proc.\@ of the 5th \\ Workshop of the ACL Special
          Interest Group in Computational Phonology (SIGPHON)}, pp.\@
        46-56, \\ Luxembourg, Aug.\@ 2000.
       \end{flushright}}} \\
       Multi-Syllable Phonotactic Modelling}
\author{Anja Belz\\
CCSRC, SRI International\\
23 Millers Yard, Mill Lane\\
Cambridge CB2 1RQ, UK\\
{\tt anjab@cam.sri.com}}

\begin{document}
\thispagestyle{plain}

\maketitle

\begin{abstract}

This paper describes a novel approach to constructing phonotactic
models. The underlying theoretical approach to phonological
description is the multi-syllable approach in which multiple syllable
classes are defined that reflect phonotactically idiosyncratic
syllable subcategories. A new finite-state formalism, {\sc ofs}
Modelling, is used as a tool for encoding, automatically constructing
and generalising phonotactic descriptions.  Language-independent
prototype models are constructed which are instantiated on the basis
of data sets of phonological strings, and generalised with a
clustering algorithm.  The resulting approach enables the automatic
construction of phonotactic models that encode arbitrarily close
approximations of a language's set of attested phonological forms.
The approach is applied to the construction of multi-syllable
word-level phonotactic models for German, English and Dutch.

\end{abstract}

\newcommand{\NLP}{\textsc{nlp }}
\newcommand{\NL}{\textsc{nl }}
\newcommand{\NLs}{\textsc{nl}s }
\newcommand{\DFA}{\textsc{dfa }}
\newcommand{\DFAn}{\textsc{dfa}}
\newcommand{\DFAs}{\textsc{dfa}s }
\newcommand{\FSA}{\textsc{fsa }}
\newcommand{\FSAs}{\textsc{fsa}s }
\newcommand{\TTS}{\textsc{tts }}
\newcommand{\GI}{\textsc{gi }}
\newcommand{\RHS}{\textsc{rhs }}
\newcommand{\RHSs}{\textsc{rhs}s }
\newcommand{\LHS}{\textsc{lhs} }
\newcommand{\LHSs}{\textsc{lhs}s }
\newcommand{\AI}{\textsc{ai} }
\newcommand{\FAn}{\textsc{fa}}
\newcommand{\FA}{\textsc{fa} }
\newcommand{\FAs}{\textsc{fa}s }
\newcommand{\OFS}{\textsc{ofs} }

\newcounter{def}
\addtocounter{def}{1}
\newenvironment{definition}
  {\begin{quote}
     \line(1,0){190}\\
     {\bf Definition \thedef}
     \addtocounter{def}{1} }
{\line(1,0){190} \end{quote} }

\section{Introduction}

Finite-state models of phonotactics have been used in automatic
language identification \cite{zissi95,belza00}, in speech recognition
\cite{carsj93,jusea94,jusea96,carsj00}, and optical character
recognition, among other applications.  While statistical models
($n$-gram or Markov models) are derived automatically from data, their
symbolic equivalents are usually constructed in a painstaking manual
process, and --- because based on standard single-syllable
phonological analyses --- tend to overgeneralise greatly over a
language's set of wellformed phonological strings.  This paper
describes methods that enable the automatic construction of symbolic
phonotactic models that are more accurate representations of
phonological grammars.

The underlying theoretical approach to phonological description is the
{\it Multi-Syllable Approach}\/ \cite{belza98d,belza00}. Syllable
phonotactics vary considerably not only in correlation with a
syllable's position within a word, but also with other factors such as
position relative to word stress.  Analyses based on multiple syllable
classes defined to reflect such factors can more accurately account
for the phonologies of natural languages than analyses based on a
single syllable class.

{\it Object-Based Finite State Modelling}\/ (previously described in
Belz, 2000) is used as an encoding, construction and generalisation
tool, and facilitates {\it Language-Independent Prototyping}\/, where
incompletely specified generic models are constructed for groups of
languages and subsequently instantiated and generalised automatically
to fully specified, language-specific models using data sets of
phoneme strings from individual languages. The theory-driven (manual)
component in this construction method is restricted to specifying the
maximum possible ways in which syllable phonotactics may differ in a
family of languages, without hardwiring the differences into the final
models. The actual construction of models for individual languages is
a data-driven process and is done automatically.

Sets of German, English and Dutch syllables were used extensively in
the research described in this paper, both as a source of evidence in
support of the multi-syllable approach (Section~\ref{sec:MSH}) and as
data in automatic phonotactic model construction
(Section~\ref{sec:models}).  All syllable sets were derived from sets
of fully syllabified, phonetically transcribed forms collected from
the lexical database {\sc celex} \cite{baayr95}.  {\sc celex}
contains compounds and phrases as well as single words.  Phonological
words were defined as any phonetic sequence with a single primary
stress marker, and all other entries were disregarded.

\begin{table*}[!thb]
\begin{center}
\begin{tabular}{|l|rr|rr|rr|}
\hline
         &\multicolumn{2}{c|}{German}&\multicolumn{2}{c|}{English}&\multicolumn{2}{c|}{Dutch}\\
         &    all &\multicolumn{1}{r|}{unique (\%)}&    all &\multicolumn{1}{r|}{unique (\%)}&    all &\multicolumn{1}{r|}{unique (\%)}\\
\hline
Initial       & 3,806  & 624 (16.4\%)    & 6,177  & 2,657 (43.01\%)  & 5,476  & 947 (17.29\%) \\
Medial        & 3,832  & 358 (9.34\%)    & 3,149  & 344 (10.92\%)    & 5,446  & 723 (13.28\%) \\
Final         & 7,040  & 2,133 (30.3\%)  & 6,750  & 2,132 (31.59\%)  & 7,279  & 1,786 (24.54\%) \\
Monosyllables & 5,114  & 855 (16.72\%)   & 7,265  & 2,963 (40.78\%)  & 5,641  & 718 (12.73\%)   \\
\hline
TOTAL         & 10,606 & 3,970 (37.43\%) & 14,333 & 8,096 (56.49\%)  & 11,448 & 4,174 (36.46\%) \\
\hline
\end{tabular}
\caption{Syllable set sizes and number of syllables unique to each set (position).}
\label{tab:syllable_stats_pos}
\end{center}
\end{table*}

\section{Multi-Syllable Phonotactics}\label{sec:MSH}

The multi-syllable approach works on the assumption that
single-syllable approaches cannot adequately capture the phonological
grammars of natural languages, because they fail to account for the
significant syllable-based phonotactic variation resulting from a
range of factors that is evident in natural languages, and
consequently overgeneralise greatly.


\paragraph{Single-syllable analyses.}  The traditional 
view is that all syllables in a language share the same structure and
compositional constraints which can be captured by a single
analysis. In many languages, however, the sets of word-initial and/or
word-final consonant clusters differ significantly from other
consonantal clusters \cite[p.\ 107ff, lists several examples from
different languages]{goldj90}. Such idiosyncratic clusters have been
treated as `terminations', `appendices', or as `extrasyllabic'
\cite{goldj90}, and integrated along with syllables at the word-level.
Similar, apparently irregular phenomena occur in correlation with tone
and stress, and the first and last vocalic segments in phonological
words are often analysed as `extratonal' and `extrametrical'.
However, such apparent irregularities are not restricted to the
beginnings and ends of phonological words, and the phonotactics of
syllables are affected by a range of factors other than position,
which are difficult if not impossible to account for by the notion of
extrasyllabicity.

Three problematic issues arise in single-syllable analyses.  Firstly,
if a phonotactic model assumes a single syllable class for a language,
and if the language has idiosyncratic word-initial and word-final
phonotactics, then the set of possible phonological words that the
model encodes is necessarily too large, and includes words that form
systematic (rather than accidental) gaps in the languages. Secondly,
if extrasyllabicity is used to account for phonotactic idiosyncracies,
then the resulting theory of syllable structure fails to account for
everything that it is intended to account for, and is forced to
integrate constituents that are not syllables (the {\it
extra}\/syllabic material) at the word level. Thirdly, the notion of
extrasyllabicity only works for cases where phonemic material can be
segmented off adjacent syllables (most easily done at the beginnings
and ends of words), and cannot be used to account for
syllable-internal variation.  The alternative offered by
multi-syllable analyses is to make the universal assumption that
position, stress and tone (among other factors) will result in
variation in syllable phonotactics that are not necessarily restricted
to any particular part of words, and to account for such variation
systematically by the use of different syllable classes.

\begin{table*}[!thb]
\begin{center}
\begin{tabular}{|l|l|ccc|}
\hline
         &         & Medial       & Final        & Mono\\
\hline
         & Initial & 2,619 (0.52) & 1,466 (0.16) & 1,392 (0.18)\\
German:  & Medial  &              & 1,928 (0.22) & 1,185 (0.15)\\
         & Final   &              &              & 3,873 (0.47)\\
\hline
         & Initial & 1,860 (0.25) & 1,920 (0.17) & 2,266 (0.20)\\
English: & Medial  &              & 1,787 (0.22) & 1,008 (0.11)\\
         & Final   &              &              & 3,576 (0.34)\\
\hline
         & Initial & 3,594 (0.49) & 2,764 (0.28) & 3,003 (0.37)\\
Dutch:   & Medial  &              & 3,279 (0.35) & 2,428 (0.28)\\
         & Final   &              &              & 4,320 (0.50)\\
\hline
\end{tabular}
\caption{Intersections and set similarities for German, English and Dutch syllables (position).}
\label{tab:GED-intersections}
\end{center}
\end{table*}

\begin{table*}[!htb]
\begin{center}
\begin{tabular}{|l|rr|rr|rr|}
\hline
         &\multicolumn{2}{c|}{German}&\multicolumn{2}{c|}{English}&\multicolumn{2}{c|}{Dutch}\\
         &    all &\multicolumn{1}{r|}{unique (\%)}&    all &\multicolumn{1}{r|}{unique (\%)}&    all &\multicolumn{1}{r|}{unique (\%)}\\
\hline
Stressed &  8,919 & 2,977 (33.37\%) &  9,399 & 5,280 (56.18\%) &  9,934 & 3,484 (35.07\%)\\
Pretonic &    989 &    30 (3.03\%)  &  3,201 & 1,362 (42.55\%) &  1,780 &    71 (3.99\%)\\
Posttonic&  5,897 &   388 (6.58\%)  &  4,754 &   670 (14.09\%) &  5,960 &   517 (8.67\%)\\
Plain    &  6,819 &   229 (3.36\%)  &  6,020 &   944 (15.68\%) &  6,662 &   176 (2.64\%)\\
\hline
TOTAL    & 10,598 & 3,624 (34.20\%) & 14,333 & 8,256 (57.60\%) & 11,443 & 4,248 (37.12\%)\\
\hline
\end{tabular}
\end{center}
\caption{Syllable set sizes and number of syllables unique to each set (stress).}
\label{tab:syllable_stats_stress}
\end{table*}

\paragraph{Related approaches.}

The idea to discriminate between different syllable types, classified
by word position and position with respect to the stressed syllable
has been explored and utilised in previous research, for example in
{\sc fsa}-based phonotactic models, typed formalisms, and in
stochastic production rule grammars. Carson-Berndsen
\shortcite{carsj93} uses two separate \FSAs to encode the phonotactics
of full and reduced syllables, and Jusek et al.\ \shortcite{jusea94}
distinguish between stressed and unstressed syllables.  In a typed
feature system of morpho-phonology, Mastroianni and Carpenter
\shortcite{mastm94} define subtypes of the general type $syllable$.

The most closely related existing research is that presented by
Coleman and Pierrehumbert \shortcite{colej97}.  The paper examines
different possibilities for using a probabilistic grammar for English
words to model native speakers' acceptability judgments.  The
production rule grammar encodes the phonotactics of English
monosyllabic and bisyllabic words.  Different probability
distributions over paths in derivation trees are investigated which
model likelihood of acceptability to native speakers, rather than
likelihood of occurrence. To build a grammar that accounts for
interactions among onsets and rhymes, location with respect to the
word edge and word stress patterns, six syllable types are
distinguished which reflect possible combinations of the features
strong, weak, initial and final.  The subsyllabic constituents onset
and rhyme are similarly marked for stress and position.

The present research extends existing work on syllable subclasses by
applying the multi-syllable approach systematically to model the
entire phonotactics of languages, and by using it for
language-independent prototyping (see
Section~\ref{sec:Lang-Indep-Phon} below).

\paragraph{Position-correlated phonotactic variation.}  
Table~\ref{tab:syllable_stats_pos} shows statistics for sets of
monosyllabic words and initial, medial and final syllables in {\sc
celex}.  For each language and each syllable set, the table shows the
size of the set (e.g.\ there are $3,806$ different initial German
syllables in {\sc celex}), and the size of its subset of syllables
that do not occur in any other set (e.g.\ $624$ out of $3,806$ initial
German syllables, or 16.4\%, only occur word-initially). For all
three languages, the figures show significant differences between the
sets of syllables that can occur in the four different positions and
their unique subsets. In German and Dutch, final syllables are
particularly idiosyncratic, with $30.3$\% and $24.54$\%, respectively,
not occurring in any other position.  In English, all syllable sets
except the medial syllables display a high degree of
idiosyncracy. Table~\ref{tab:GED-intersections} shows the size of the
intersections between the syllable sets, and the more objective
measure of set similarity in brackets\footnote{Set similarity here is
the standard measure of the size of the intersection over the size of
the union of two sets $S_{1}$ and $S_{2}$, or $|S_{1} \cap
S_{2}|/|S_{1} \cup S_{2}|$ (not defined for $S_{1} = S_{2} =
\emptyset$).}.  In German and Dutch, the similarity between initial
and medial syllables, and between final and monosyllables is
particularly high.  The similarity between the least similar of
syllable sets is much greater in Dutch than in either English or
German. In English, only the final and monosyllables display any
significant similarity.  Average set similarity is highest in Dutch
(0.37), followed by German (0.28), and English (0.21).

\paragraph{Stress-correlated phonotactic variation.} 
Table~\ref{tab:syllable_stats_stress} shows analogous statistics for
phonotactic variation correlated with word stress.  Set sizes and
unique subset sizes are shown for the set of syllables that carry
primary stress (stressed), those immediately preceding stress
(pretonic), those immediately following stress (posttonic), and all
others (plain).  In all three languages, the set of stressed syllables
has least in common with other sets.  In English, this is closely
followed by the pretonic syllables.  The average percentage of
syllables unique to a set is highest in English, followed by Dutch and
then German.

\paragraph{}
These statistics show not only that there is significant
syllable-level variation in the phonotactics of all three languages,
but also that the simple strategy of subdividing the set of all
syllables on the basis of position and stress succeeds in capturing at
least some of this variation.  If a high percentage of syllables in
one subcategory do not occur in any other, then distinguishing this
syllable subcategory in a phonotactic model will help reduce
overgeneralisation.

\section{Encoding, Construction and Generalisation of Phonotactic Models}

\subsection{Object-Based Finite-State Modelling}

The \OFS Modelling formalism was used as a tool for encoding,
constructing and generalising phonotactic models in the research
described in Section~4.  \OFS Modelling consists of three main
components, (i) a representation formalism, (ii) a mechanism for
automatic model construction, and (iii) mechanisms for model
generalisation.  Brief summaries of the components that were used in
the research described in this paper are given here (for full details
see Belz, 2000).

Underlying \OFS Modelling is a set of assumptions about linguistic
description that shares many of the fundamental tenets of declarative
phonology \cite[for example]{birds91}. This set of assumptions
includes a strictly non-derivational, non-transformational and
constraint-based approach to linguistic description, and the principle
of constraint inviolability.

The \OFS formalism is a declarative, monostratal finite-state
representation formalism that is intuitively readable, facilitates the
automatic data-driven construction of models, and permits the
integration of available prior, theoretical knowledge.  The
derivations (trees or brackettings) defined by \OFS models correspond
to context-free derivations with a limited tree depth or degree of
nesting of brackets.  This means that in \OFS models (unlike in other
normal forms for regular grammars), rules (hence expansions or
brackets) can, if appropriately defined, systematically correspond to
standard linguistic objects, the reason why the formalism is called
{\it object-based}\/.

\begin{figure}[!h]
\begin{center}
\begin{tabular}{llll}
\hline
\multicolumn{4}{l}{\OFS Model $O = ( N, T, P, n+1) $} \\
\hline
\vspace{0.2cm}
n:   & $O^{n}_0$          & $\Rightarrow$ & $\omega^{n}_0$ \\
\vspace{0.2cm}
n-1: & $O^{n-1}_{0}$    & $\Rightarrow$ & $\omega^{n-1}_{0}$ \\
\vspace{0.2cm}
     & $O^{n-1}_{1}$    & $\Rightarrow$ & $\omega^{n-1}_{1}$ \\
     & $\cdots$ & & \\
     & $O^{n-1}_{m}$    & $\Rightarrow$ & $\omega^{n-1}_{m}$ \\
$\ldots$ & & & \\
\vspace{0.2cm}
1:   & $O^{1}_{0}$      & $\Rightarrow$ & $\omega^{1}_{0}$ \\
\vspace{0.2cm}
     & $O^{1}_{1}$      & $\Rightarrow$ & $\omega^{1}_{1}$ \\
     & $\cdots$ & & \\
\vspace{0.2cm}
     & $O^{1}_{l}$      & $\Rightarrow$ & $\omega^{1}_{l}$ \\
\vspace{0.2cm}
0:   & $O^{0}_{0}$      & $\Rightarrow$ & $\omega^{0}_{0}$ \\
\vspace{0.2cm}
     & $O^{0}_{1}$      & $\Rightarrow$ & $\omega^{0}_{1}$ \\
     & $\cdots$ & & \\
\vspace{0.2cm}
     & $O^{0}_{p}$      & $\Rightarrow$ & $\omega^{0}_{p}$ \\
\hline
\end{tabular}
\end{center}
\caption{Notational convention for \OFS models.}
\label{fig:gen-ofs-model}
\end{figure}

\paragraph{OFS Models.}  The \OFS representation formalism is essentially a 
normal form for regular sets.  \OFS models can be interpreted in the
same way as standard production rule grammars, but are subject to a
set of additional constraints.  An \OFS model $O$ is denoted $(N, T,
P, n+1)$, where $N$ is a finite set of non-terminal objects $O^{i}_j$,
$0 \leq i \leq n$, and $T$ is a finite set of terminals.  $P$ is an
ordered finite set of $n$ sets of productions $O^{i}_j \Rightarrow
\omega^i_j$, where $O^{i}_j \in N$, and for $i > 0$, $\omega^i_j$ is a
regular expression\footnote{In the regular expressions in this paper,
$r^*$ denotes any number of repetitions of $r$, $r^+$ denotes at least
one repetition of $r$, and $r + e$ denotes the disjunction of $r$ and
$e$.} over symbols $O^{g}_h \in N, i > g$, whereas for $i = 0$,
$\omega^i_j$ is a set of strings\footnote{The string sets in level~0
\RHSs are actually implemented more efficiently as finite automata.}
from $T^{*}$. An \OFS model $O$ has $n$ levels, or sets of production
rules, and each rule $O^{i}_j \Rightarrow \omega^i_j$ is uniquely
associated with one of the levels.  The $n$th set of production rules
is a singleton set $\{ O^{n}_0 \Rightarrow \omega^n_0 \}$, and $O^{n}_0$ is
interpreted as the start symbol.  The notational convention adopted
for \OFS models is as shown in Figure~\ref{fig:gen-ofs-model}.

\begin{small}
\begin{definition} {\it OFS Model}

An \OFS model $O$ is a 4-tuple $(N, T, P, n~+~1)$, where $N$ is a
finite set of nonterminals $O^{i}_j$, $0 \leq i \leq n$, $O^{n}_0 \in N$
is the start symbol, $T$ is a finite set of terminals, $n+1$ denotes
the number of levels in the model, and $P = $

\vspace{0.2cm}
\hspace{-0.3cm}\begin{tabular}{l}
$\big\{$ $\{ O^{n}_0 \Rightarrow \omega^{n}_0 \},$\\
\vspace{0.2cm}
\hspace{.17cm} $\{ O^{n-1}_{0} \hspace{-.1cm} \Rightarrow \hspace{-.1cm} \omega^{n-1}_{0}, $ $ O^{n-1}_{1} \hspace{-.1cm} \Rightarrow \hspace{-.1cm} \omega^{n-1}_{1}, $ $ \ldots O^{n-1}_{m} \hspace{-.1cm} \Rightarrow \hspace{-.1cm} \omega^{n-1}_{m} \},$ \\
\vspace{0.2cm}
\hspace{.17cm} $\ldots$ \\
\vspace{0.2cm}
\hspace{.17cm} $\{ O^{1}_{0} \Rightarrow \omega^{1}_{0}, $ $ O^{1}_{1} \Rightarrow \omega^{1}_{1}, $ $ \ldots O^{1}_{l} \Rightarrow \omega^{1}_{l} \},$ \\
\vspace{0.2cm}
\hspace{.17cm} $\{ O^{0}_{0} \Rightarrow \omega^{0}_{0}, $ $  O^{0}_{1} \Rightarrow \omega^{0}_{1}, $ $ \ldots O^{0}_{p} \Rightarrow \omega^{0}_{p}$ $\} $\hspace{.1cm} $  \big\} $, \\
\end{tabular}

where each rule $O^{i}_j \Rightarrow \omega^i_j$ is uniquely
associated with one of the levels, $\omega^0_j$ is a set of strings
from $T^*$, $\omega^i_j, i > 0$, is a regular expression over objects
$O^g_h \in N, i > g$.

\end{definition}
\end{small}

Each rule $O \Rightarrow \omega$ in an \OFS model corresponds to a set
of strings which will be referred to as an object set or class, where
$O$ is the name of the object.  The production rules in \OFS models
will also be referred to as object rules.

\OFS models thus differ from standard production rule grammars in
three ways. Firstly, \RHSs of rules above level~0 are arbitrary
regular expressions\footnote{Other formalisms for linguistic analysis
have permitted full regular expressions in the \RHSs of rules.  For
instance, in syntactic grammars, the recursive nature of some types of
coordination has been modelled with right-recursive regular
expressions (e.g.\ in {\sc gpsg}).}.  Secondly, terminals from $T$ are
restricted to appearing in the \RHSs of rules at level~0 (mostly to
facilitate automatic model construction, see below). Thirdly, \OFS
models are limited in their representational power to the finite-state
domain by the constraints that the \RHSs of rules in rule sets at
level $i > 0$ are regular expressions over non-terminals that appear
only in the \LHSs of rules in rule sets at levels $g < i$.  That this
limits representational power to the regular languages can be seen
from the fact that all non-terminals $O^i_j$ in the \RHS of the single
top-level rule can be substituted iteratively with the \RHSs of the
corresponding rules $O^i_j \Rightarrow \omega^i_j$.  This iteration
terminates after a finite time because there is a finite number of
levels in the model, and at this point the \RHS of the top-level rule
contains only non-terminals, i.e.\ $is$ a regular expression, hence
represents a regular language.

Unlike other normal forms for regular production-rule grammars (such
as left-linear and right-linear sets of production rules), \OFS models
enable the definition of production rules and hence derivations that
can, if appropriately defined, correspond to standard linguistic
objects and constituents (not possible in linear grammars).  Through
the association of rules with a finite number of levels, \OFS models
permit the definition of grammars that encode sets of context-free
derivations up to a maximum depth equal to the number of levels in the
model.

The fact that non-terminal strings are in \OFS models restricted to
the lowest level, facilitates the combined theory and data driven
construction of models.  Uninstantiated models can be defined, that
encode what is known in advance about the structural regularities of
the object to be modelled in levels above 0, and have under-specified
level~0 \RHSs that are subsequently instantiated on the basis of data
sets of examples of the object to be modelled.  \OFS Modelling also
has a generalisation procedure which can be used to generalise fully
instantiated \OFS models.  Each of these mechanisms is described in
turn over the following paragraphs.

\paragraph{Uninstantiated OFS Models.}  In fully specified \OFS models (as
defined in the preceding section), the right-hand sides ({\sc rhs}s)
of production rules at level $i$ are regular expressions for $i > 0$,
and string sets for $i = 0$.  This separation makes it simple to
construct incompletely specified models, or {\it prototype OFS
models}, where the \RHSs of level~0 rules are pattern descriptions
rather than strings sets.  Level~0 \RHSs in prototype models have the
form $O^{0}_{i} \Rightarrow S_{i}$, where $O^{0}_{i}$ is the name of
the object, and $S_{i}$ is a set former $\{ x : {\mathbf v}x{\mathbf
w} \in D, P_{1}, P_{2}, \ldots P_{n} \}$, where ${\mathbf v}, {\mathbf
w}$ are concatenations of variables, $D$ refers to any given finite
data set of strings, and $P_{i}, 1 \leq i \leq n$ are properties of
the variables in ${\mathbf v}$ and ${\mathbf w}$.

\paragraph{Instantiation of Prototype OFS Models.}  The \OFS 
instantiation procedure takes a prototype \OFS model $M$ for some
linguistic object and a data set $D$ of example members of the
corresponding object class and proceeds as follows.  For each level~0
rule $O^{0}_{i} \Rightarrow S_{i}$ in $M$, and for each element $x$ of
$D$, all substrings of $x$ that match $S_{i}$ are collected.  The
resulting set of substrings becomes the new \RHS of rule
$O^{0}_{i}$. After instantiation, level~0 rules whose \RHS is the
empty set are removed, as are rules at higher levels whose \RHSs
contain non-terminals that can no longer be expanded by any of the
production rules in $M$.

\begin{figure*}[!hbt]
\begin{center}
\begin{tabular}{llll}
\hline
\multicolumn{4}{l}{Prototype \OFS Model $Syllable = ( \{ Syllable, Onset,
Peak, Coda \}, T, P, 2 )$}\\
\hline
1: & $Syllable$  & $\Rightarrow$ & $Onset \hspace{.15cm} Peak \hspace{.15cm} Coda$ \\
0: & $Onset$     & $\Rightarrow$ & $\{x\hspace{.1cm} |\hspace{.1cm} xay \in D, x \in$ {\it\small CONSONANTS}$^{*}, a \in$ {\it\small VOWELS}$ \} $\\
   & $Peak$      & $\Rightarrow$ & $\{x\hspace{.1cm} |\hspace{.1cm} yxz \in D, x \in$ {\it\small VOWELS}$^{+}, y,z \in$ {\it\small CONSONANTS}$^{*} \} $\\
   & $Coda$      & $\Rightarrow$ & $\{x\hspace{.1cm} |\hspace{.1cm} yax \in D, x \in$ {\it\small CONSONANTS}$^{*}, a \in$ {\it\small VOWELS} $\} $\\
\hline
\end{tabular}
\end{center}
\caption{Simple prototype \OFS model for syllable-level phonotactics.}
\label{fig:proto-grapho-model}
\end{figure*}

\begin{figure*}[!htb]
\begin{tt}
\begin{center}
\begin{tabular}{|l|}
\hline

\begin{IPA}
{\ae}z, {\ae}S, A:sk, {\ae}sp, {\ae}s, {\ae}t, Et, O:k, O:ks,
A:nts, O:, O:z, {\ae}ks, aI, aIz, beI, bA:, bA:z, beIb,
b{\ae}k, \end{IPA}\\

\begin{IPA} 
b{\ae}ks, si:, k{\ae}b, {\textteshlig}E@$\ast$,
{\textteshlig}E@d, sIn{\textteshlig}, sIn{\textteshlig}t, kli:v,
dEf, di:l, dju:st, d2vz, drA:fts, dwEld, faI, frEt, \end{IPA}\\

\begin{IPA}
g@Uld, gr6t, kwId, spl{\ae}t, sprIN, str{\ae}ps, st2n
\end{IPA} \\

\hline
\end{tabular}
\end{center}
\end{tt}
\caption{Small data set of English monosyllabic words.}
\label{fig:toy-text-data-set}
\end{figure*}

\begin{figure*}[!htb]
\begin{center}
\begin{tabular}{llll}
\hline
\multicolumn{4}{l}{\OFS Model $Syllable = ( \{ Syllable, Onset,
Peak, Coda \}, T, P, 2 )$}\\
\hline
1: & $Syllable$  & $\Rightarrow$ & $Onset \hspace{.15cm} Peak \hspace{.15cm} Coda$ \\
0: & $Onset$     & $\Rightarrow$ & $\{$ $\epsilon$, \begin{IPA} b, s, k, st, f, d, {\textteshlig}, kl, dj, dr, dw, fr, g, gr, kw, spl, spr, str\end{IPA} $\}$\\
   & $Peak$      & $\Rightarrow$ & $\{$\begin{IPA}{\ae}, A:, E, O:, aI, eI, i:, E@, 2, @U, 6, I, u: \end{IPA} $ \}$\\
   & $Coda$      & $\Rightarrow$ & $\{$ $\epsilon$, \begin{IPA} b, s, k, st, f, d, z, S, sk, sp, ks, nts, $\ast$, n{\textteshlig}, n{\textteshlig}t, v, l, vz, fts, ld, t, N, ps, n \end{IPA} $ \}$\\
\hline
\end{tabular}
\end{center}
\caption{Syllable-level phonotactic \OFS model instantiated with set of English monosyllables.}
\label{fig:grapho-model}
\end{figure*}

\begin{figure*}[!htb]
\begin{center}
\begin{tabular}{llll}
\hline
\multicolumn{4}{l}{\OFS Model $Syllable = ( \{ Syllable,
Onset\_Coda, Peak, \}, T, P, 2 )$}\\
\hline
1: & $Syllable$    & $\Rightarrow$ & $Onset\_Coda \hspace{.15cm} Peak \hspace{.15cm} Onset\_Coda$ \\
0: & $Onset\_Coda$ & $\Rightarrow$ & $\{$ $\epsilon$, \begin{IPA} b, s, k, st, f, d, {\textteshlig}, kl, dj, dr, dw, fr, g, gr, kw, spl, spr, str,\end{IPA}\\
   &               &               & \hspace{.2cm} \begin{IPA}z, S, sk, sp, ks, nts, $\ast$, n{\textteshlig}, n{\textteshlig}t, v, l, vz, fts, ld, t, N, ps, n \end{IPA} $\}$\\
   & $Peak$        & $\Rightarrow$ & $\{$\begin{IPA}{\ae}, A:, E, O:, aI, eI, i:, E@, 2, @U, 6, I, u: \end{IPA} $\}$\\
\hline
\end{tabular}
\end{center}
\caption{\OFS model of Figure~\ref{fig:grapho-model} generalised with $\tau \leq 0.19$.}
\label{fig:generalised-grapho-model}
\end{figure*}

\paragraph{Object-Set Generalisation.}  Instantiated \OFS models can be
generalised by object-set ({\sc os}) generalisation, where pairs of
level~0 object sets are compared on the basis of a standard
set similarity measure $sim$ for two finite sets $D_{1}$ and $D_{2}$
(not defined for $D_{1} = D_{2} = \emptyset$): $sim(D_{1},D_{2}) =
|D_{1} \cap D_{2}| / |D_{1} \cup D_{2}|$.  The {\sc os}-generalisation
procedure takes a fully specified \OFS model $M$ and a given
similarity threshold $\tau$, and, applying a simple clustering
algorithm, merges all object sets that have a similarity value $sim$
matching or exceeding $\tau$.  That is, the {\sc os}-generalisation
procedure measures the similarity between all pairs of level~0 sets,
and all pairs that match or exceed the threshold end up in the same
cluster.  Finally, the old object names (non-terminals) in the \RHSs
of object rules at levels above~0 are replaced with the \LHSs of the
corresponding new merged object rule, while all object rules that now
have identical \RHSs are in turn merged. In this way, generalisation
`percolates' upwards through the levels of the model.

Determining an appropriate value for the similarity threshold $\tau$
is not unproblematic.  It could be set in relation to the average
similarity value in an instantiated model (individually for each
prototype instantiation), but this approach would obscure the
similarities that object-set generalisation (in particular in
conjunction with {\sc lip}) is intended to exploit.  The whole point
of object-set generalisation for language-independent prototypes is
that it will merge a different number of level~0 object classes in
different prototype instantiations, creating different final,
language-specific \OFS models. If $\tau$ is set in proportion to the
average similarity between level~0 classes, then this difference is
reduced, and the resulting models will tend to retain the same number
of level~0 object classes from the prototype.  For example, if the
above prototype model $Word$ is instantiated to a data set from a
language that has phonotactics which differ only between stressed and
unstressed syllables, then all similarity values between stressed
syllable classes regardless of their position within a word, and
between all posttonic, pretonic and plain syllables classes (again,
regardless of position), will be very high. The average similarity
value will therefore also be high.  If $\tau$ is set in relation to
this high average, not all unstressed and all stressed syllable
classes, respectively, will be merged, because not all syllable
classes can exceed average similarity.

Average similarity is a language-specific property, and so is the
number of syllable classes similar enough to be merged for a given
$\tau$ value.  For different generalised instantiations of the same
prototype model to be comparable, object-set generalisation must have
been carried out for each of them with the same $\tau$ value.

The threshold $\tau$ is best regarded as a variable parameter to the
{\sc os}-generalisation procedure that can be used to control the
degree to which a generalised {\sc ofs} model will fit the data: the
higher $\tau$, the more closely the model will fit the data, and the
less it will generalise over it.  This is particularly appropriate in
phonotactic modelling, because phonotactics seeks to encode not just
the set of attested words, but also unattested, but wellformed words
(often called `accidental' gaps), while excluding only illformed words
(or `systematic' gaps).  There is no objective dividing line between
idiosyncratic and systematic gaps, and setting $\tau$ can be used as
one way of controlling the degree of conservativeness in generalising
over the set of attested words.

\subsection{Example}

As an illustration, consider the following example construction of a
simple \OFS model for syllable-level phonotactics (the constraints
that hold on the possible phoneme sequences within
syllables)\footnote{The example model is not intended to be a
realistic phonotactic model, but is provided here merely as an
illustration of the techniques outlined above.}.  The prototype \OFS
model constructed in the first step
(Figure~\ref{fig:proto-grapho-model}) encodes the standard assumption
that the syllable-level phonotactics in different languages can be
appropriately modelled by interpreting syllables as a sequence of
consonantal phonemes (onset), followed by a sequence of vocalic
phonemes (peak), and another sequence of consonantal phonemes (coda).

In the second construction step, a data set of English monosyllabic
words (Figure~\ref{fig:toy-text-data-set}) is used to instantiate the
prototype \OFS model.  The instantiation procedure constructs an \OFS
model with new level~0 \RHSs as shown in
Figure~\ref{fig:grapho-model}. During {\sc os}-generalisation, $sim$
values are computed for each pair of level~0 object sets.  The only
pairwise intersection that is non-empty (hence the only non-zero $sim$
value) in this example is that between the sets $Coda$ and $Onset$
($sim = 0.19$), which are merged if {\sc os}-generalisation is applied
to \OFS model $Syllable$ with $\tau \leq 0.19$, resulting in the
simpler, more general \OFS model shown in
Figure~\ref{fig:generalised-grapho-model}.

\begin{figure*}[!ht]
\begin{center}
\begin{tabular}{llll}
\hline
\multicolumn{4}{l}{Prototype {\sc ofs} Model $Word = (N, M, P, 2) $} \\
\hline
1: & $Word$ & $\Rightarrow$ & $S\_mon\_st \hspace{.15cm}$  +  \\
   &        &               & $S\_mon\_pl \hspace{.15cm}$  +  \\
   &        &               & ($S\_ini\_st$  $S\_fin\_po) \hspace{.15cm}$  +  \\
   &        &               & ($S\_ini\_st$  $S\_med\_po$  $S\_med\_pl^{*}$ $S\_fin\_pl) \hspace{.15cm}$  +  \\
   &        &               & ($S\_ini\_pr$ $S\_fin\_st) \hspace{.15cm}$  +  \\
   &        &               & ($S\_ini\_pr$ $S\_med\_st$ $S\_fin\_po) \hspace{.15cm}$ + \\
   &        &               & ($S\_ini\_pr$  $S\_med\_st$ $S\_med\_po$ $S\_med\_pl^{*}$ $S\_fin\_pl) \hspace{.15cm}$   +              \\
   &        &               & ($S\_ini\_pl$ $S\_med\_pl^{*}$ $S\_med\_pr$ $S\_fin\_st) \hspace{.15cm}$  + \\
   &        &               & ($S\_ini\_pl$ $S\_med\_pl^{*}$ $S\_med\_pr$ $S\_med\_st \hspace{.15cm}$ $S\_fin\_po) \hspace{.15cm}$ + \\
   &        &               & ($S\_ini\_pl$ $S\_med\_pl^{*}$ $S\_med\_pr$  $S\_med\_st$ $S\_med\_po$ $S\_med\_pl^{*}$ $S\_fin\_pl$)                                  \\
0: & $S\_mon\_st$ & $\Rightarrow$ & $\{ x :$ $'x \in D, x \in (M \backslash \{ - \})^{*} \} $  \\
   & $S\_mon\_pl$ & $\Rightarrow$ & $\{ x : x \in D, x \in (M \backslash \{ -, ' \})^{*} \} $  \\
   & $S\_ini\_st$ & $\Rightarrow$ & $\{ x :$ $'x-w \in D, x \in (M \backslash \{ - \})^{*} \} $  \\
   & $S\_ini\_pr$ & $\Rightarrow$ & $\{ x : x-'vw \in D, x,v \in (M \backslash \{ - \})^{*} \} $  \\
   & $S\_ini\_pl$ & $\Rightarrow$ & $\{ x : x-u-'vw \in D, x,v \in (M \backslash \{ - \})^{*} \} $  \\
   & $S\_med\_st$ & $\Rightarrow$ & $\{ x : v-'x-w \in D, x \in (M \backslash \{ - \})^{*} \} $  \\
   & $S\_med\_pr$ & $\Rightarrow$ & $\{ x : u-x-'vw \in D, x,v \in (M \backslash \{ - \})^{*} \} $  \\
   & $S\_med\_po$ & $\Rightarrow$ & $\{ x : u'v-x-w \in D, x,v \in (M \backslash \{ - \})^{*} \} $  \\
   & $S\_med\_pl$ & $\Rightarrow$ & $\{ x : ( u'y-v-x-w \in D ) \lor$  $( u-x-v-'w \in D ), x \in (M \backslash \{ - \} )^{*} \} $  \\
   & $S\_fin\_st$ & $\Rightarrow$ & $\{ x : w-'x \in D, x \in (M \backslash \{ - \})^{*} \} $  \\
   & $S\_fin\_po$ & $\Rightarrow$ & $\{ x : w'v-x \in D, x,v \in (M \backslash \{ - \})^{*} \} $  \\
   & $S\_fin\_pl$ & $\Rightarrow$ & $\{ x : w'v-u-x \in D, x,v \in (M \backslash \{ - \})^{*} \} $  \\
\hline
\end{tabular}
\end{center}
\caption{Prototype {\sc ofs} model for multi-syllable word-level phonotactics.}
\label{fig:ofs-model-word}
\end{figure*}

\subsection{Language-Independent Prototyping}\label{sec:Lang-Indep-Phon}

Language-independent prototyping ({\sc lip}) as a general approach to
linguistic description seeks to define generic models that restrict
--- in some linguistically meaningful way --- the set of grammars or
descriptions that can be inferred from data.  \OFS modelling can be
used as an implementational tool for {\sc lip}. Language-independent
prototype \OFS models can be defined by specifying a maximal number of
objects and corresponding production rules such that when the
prototype is instantiated and generalised with data sets from
individual languages, different object sets will be deleted and merged
for different languages, resulting in different final, instantiated
and generalised \OFS models.  In the following section, a
language-independent phonotactic prototype \OFS model is instantiated
to surprisingly different \OFS models for three closely related
languages.

\section{Multi-Syllable Phonotactic Models for German, English and Dutch}\label{sec:models}

When applied to modelling multi-syllable word-level phonotactics, {\sc
lip} with \OFS Modelling means defining the maximum possible number of
syllable classes that may be subject to different phonotactic
constraints in a given group of languages.  The exact set of syllable
classes depends on the group of languages the prototype is intended to
cover as well as the desired amount of generalisation over data (in
general, a model that distinguishes only two syllable classes will
generalise more than a model that distinguishes three or more classes,
given the same data). The prototype presented in this section is
intended to cover German, English and Dutch, and takes into account
only phonological factors (syntactic factors such as word category
which can also affect phonotactics are not taken into account).  Two
phonological factors are modelled: position of a syllable within a
word, and position of a syllable relative to primary word stress.

For this modelling task, the {\sc lip} approach is implemented by
constructing an \OFS prototype model in which syllable classes
reflecting all possible different combinations of position within a
word and relative to stress are defined as level~0 uninstantiated
object rules, and all possible ways in which the corresponding objects
can be combined to form words are defined as higher-level object
rules.  No prior assumptions about where phonotactic variation occurs
is hardwired into the model.  Instead, the maximal ways in which
phonotactics may vary in a group of languages is encoded. The idea is
that prototype instantiation and {\sc os}-generalisation with data
sets of phonological words from different languages will result in
different final, instantiated phonotactic models.

\begin{table*}[!ht]
\begin{center}
\begin{tabular}{|l|rr|rr|rr|}
\hline
&\multicolumn{2}{c|}{German}&\multicolumn{2}{c|}{English}&\multicolumn{2}{c|}{Dutch}\\
&    all &\multicolumn{1}{r|}{unique (\%)}&    all &\multicolumn{1}{r|}{unique (\%)}&    all &\multicolumn{1}{r|}{unique (\%)}\\
\hline
$Set\_mon\_st$ & 5,028 & 849 (16.89\%) & 7,254 & 2,958 (40.77\%) & 5,641 & 719 (12.75\%) \\
$Set\_mon\_pl$ & 1,813 &   1  (0.06\%) &    11 &     5 (45.45\%) &     0 &   -     (-) \\
$Set\_ini\_st$ & 3,658 & 527 (14.41\%) & 3,345 &   409 (12.23\%) & 5,258 & 772 (14.68\%) \\
$Set\_ini\_pr$ &   707 &  18  (2.55\%) & 2,560 & 1,328 (51.88\%) & 1,346 &  49  (3.64\%) \\
$Set\_ini\_pl$ & 1,628 &  19  (1.17\%) & 1,495 &   437 (29.23\%) & 1,252 &  28  (2.24\%) \\
$Set\_med\_st$ & 2,527 &  92  (3.64\%) & 1,600 &    90  (5.63\%) & 3,907 & 282  (7.22\%) \\
$Set\_med\_pr$ &   618 &  12  (1.94\%) &   916 &    30  (3.28\%) & 1,026 &  26  (2.53\%) \\
$Set\_med\_po$ & 2,518 &  66  (2.62\%) & 1,415 &    65  (4.59\%) & 3,296 & 185  (5.61\%) \\
$Set\_med\_pl$ & 2,220 &  28  (1.26\%) & 1,156 &    82  (7.09\%) & 2,897 &  36  (1.24\%) \\
$Set\_fin\_st$ & 4,261 & 822 (19.29\%) & 3,376 &   583 (17.27\%) & 4,972 & 803 (16.15\%) \\
$Set\_fin\_po$ & 4,354 & 413  (9.49\%) & 4,141 &   882  (21.3\%) & 4,525 & 460  (1.02\%) \\
$Set\_fin\_pl$ & 3,716 & 166  (4.47\%) & 2,635 &   306 (11.61\%) & 3,820 & 101  (2.64\%) \\
\hline
{\sc total}  & 10,598 & 3,013 (28.42\%) & 14,333 & 7,175 (50.06\%) & 11,443 & 3,461 (30.25\%) \\
\hline
\end{tabular}
\end{center}
\caption{Sizes of Level~0 object sets resulting from instantiations, and
syllables unique to each set.}
\label{tab:word-object-set-sizes}
\end{table*}

\subsection{Language-Independent Prototype OFS Model for Multi-syllable Phonotactics}  

The prototype model shown in Figure~\ref{fig:ofs-model-word}
distinguishes between twelve syllable classes which correspond to all
possible combinations of position within a word and position relative
to primary stress ($'$~marks primary stress, $-$ is the syllable
separator, and $S$ = syllable).  As before, the set of all syllables
is divided into four classes on the basis of position ($mon$ =
monosyllabic, $ini$ = initial, $med$ = medial, $fin$ = final), each of
which is divided further into four subclasses on the basis of stress
($st$ = stressed, $pr$ = pretonic, $po$ = posttonic, $pl$ = plain).
This results in a total of 12 possible syllable
categories\footnote{Not $4 \times 4 = 16$ classes, because some
classes cannot exist (e.g.\ there is no such thing as a posttonic
initial syllable).}. $D$ is the data set given in instantiation, and
$M$ the corresponding set of terminals (here, the phonemic symbols
that occur in $D$).  The {\sc rhs} of the level~1 object rule encodes
all possible ways in which the twelve syllable classes can
theoretically combine to form words.  The prototype model is
language-independent, because not all syllable classes will exist in
all languages (e.g.\ a language where primary stress is always on the
first syllable would not have classes of word-initial pretonic or
plain syllables), and {\sc os}-generalisation will create different
new syllable classes, depending on which classes are most similar in a
given language.

\subsection{Prototype Model Instantiations}

Table~\ref{tab:word-object-set-sizes} shows the sizes of the different
level~0 object sets resulting from \OFS model instantiations to
the German, English and Dutch word sets derived from {\sc celex} (the
syllable sets are far too large to be shown in their entirety).  In
all three languages, the largest syllable set is the set of stressed
monosyllables, and the smallest is the set of medial pretonic
syllables\footnote{Disregarding the set of plain monosyllables of
which there were no examples in the Dutch section of {\sc celex}, and
only a very small number in the English section.}.
Table~\ref{tab:word-object-set-sizes} also shows (in the same format
as in Section~2) the number of syllables in each syllable class that
do not occur in any of the other classes.

In German and Dutch, percentages of unique syllables are significantly
lower than in the classes reflecting position only and stress only
that were shown in Section~2, indicating that some of the classes may
not be worth distinguishing in phonotactic models.  In English,
however, the higher percentages of unique syllables are not far behind
those shown previously, indicating that most of the twelve syllable
classes in the prototype are worth distinguishing.

Some correlation is evident between the size of a set and the
percentage of unique syllables it contains.  In German, average
syllable set size is $2,754$ and the average percentage of unique
syllables is $6.48\%$. Five syllable sets are of above average size,
and four of these also have above-average percentages of unique
syllables.  Seven syllable sets are below average in size, and non of
these have above-average percentages of unique syllables.  In English,
the picture is not as straightforward.  Average syllable set size is
$2,717$, and average percentage of unique syllables is $18.62\%$.  Of
the four sets of above-average size, two have above-average, and two
have below-average, percentages of unique syllables.  Of the seven
English syllable sets of below-average size (the set of plain
monosyllables is disregarded again for English and Dutch), two have
above-average, and five have below-average percentages of unique
syllables.  Finally, in Dutch, average set size is $3,449$ and average
percentage of unique syllables is $6.33\%$.  Four of the six
above-average sized sets also have above-average percentages of unique
syllables, while all of the below-average sized sets also have
below-average percentages of unique syllables.  However, there is no
complete correlation, with some of the largest sets having very
small percentages of unique syllables, and vice versa.

\begin{figure*}[hbt]
\centering
\includegraphics[width=0.75\textwidth,height=0.3\textheight]{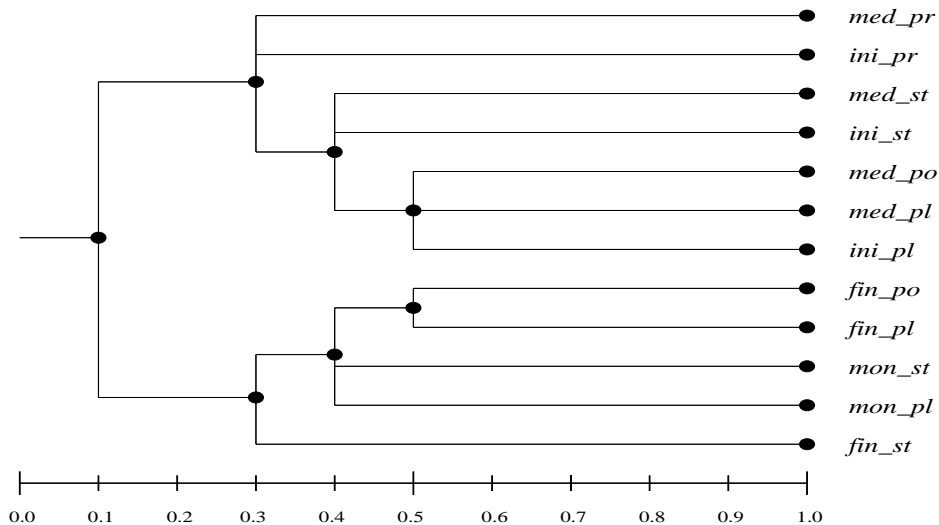}
\caption{Cluster tree for German syllable sets.}
\label{fig:german-clusters}
\end{figure*}

\subsection{OS-Generalisation of Models}

As is clear from the instantiation results presented in the preceding
section, some syllable classes contain such low percentages of unique
syllables that it is not worth distinguishing them as a separate
class.  {\sc os}-generalisation of models can be used to merge the
most similar classes and reduce the number of classes that the model
distinguishes.

\begin{figure*}[hbt]
\centering
\includegraphics[width=0.75\textwidth,height=0.3\textheight]{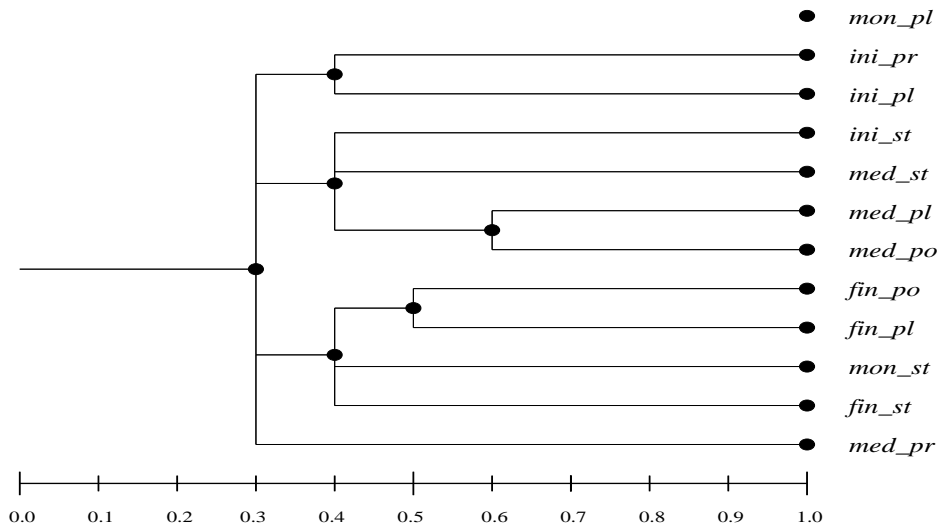}
\caption{Cluster tree for Dutch syllable sets.}
\label{fig:dutch-clusters}
\end{figure*}

\begin{figure*}[hbt]
\centering
\includegraphics[width=0.75\textwidth,height=0.3\textheight]{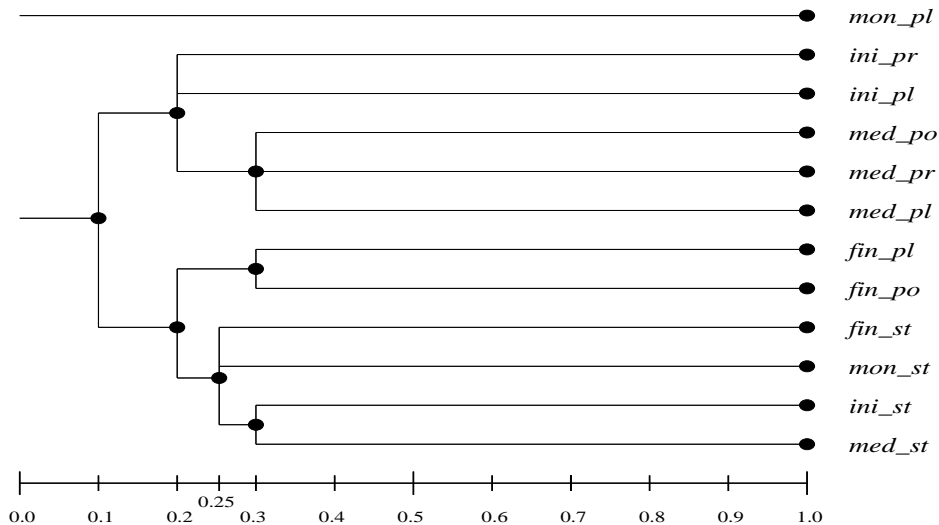}
\caption{Cluster tree for English syllable sets.}
\label{fig:english-clusters}
\end{figure*}

\subsubsection{Generalisation of Multi-Syllable OFS Model for German}\label{sec:German}

Figure~\ref{fig:german-clusters} shows the cluster tree for the German
syllable sets produced by carrying out {\sc os}-generalisation for
$\tau = 0.1 .. 1.0$ in increments of $0.1$. Each node in the tree
shows at which $\tau$ values the original syllable sets at the leaves
dominated by the node were merged. The tree reveals a very neat
picture for German.  $0.56$ is the highest $\tau$ value between any
syllable class pair, so for $\tau \geq 0.6$ no classes are
merged. $\tau = 0.5$ results in two clusters, one containing final
unstressed syllables, the other initial and medial unstressed
syllables.  At $\tau = 0.4$, all monosyllables are added to the final
syllable class, and one more medial and one more initial class to the
set of initial and medial syllables.  At $\tau = 0.3$, all
monosyllables and final syllables on the one hand, and all initial and
medial syllables on the other, are merged.  Setting $\tau$ lower makes
no difference until it is set below $0.2$, at which point all of the
original syllable classes are merged into a single set.

This shows clearly that in German the distinction between
monosyllables and final syllables on the one hand, and between initial
and medial syllables on the other, is very strongly marked
(preserved even when $\tau$ is set as low as $0.2$).  This distinction
is thus marked far more strongly than the unstressed/stressed division
(which is more commonly encoded in \DFA models of German
phonotactics), which disappears at $\tau = 0.4$ (in fact, even
earlier, at $\tau = 0.47$).

\subsubsection{Generalisation of Multi-Syllable OFS Model for Dutch}\label{sec:Dutch}

The cluster tree for Dutch (Figure~\ref{fig:dutch-clusters}) also
reveals an important division between final and monosyllables on the
one hand, and initial and medial syllables on the other. However, it
is not as clearly marked as in German. There is a point ($\tau = 0.4$)
when all final and monosyllables are in the same cluster, but this is
not the case for the initial and medial syllables, which form
subclusters that are correlated with stress.  The medial plain and
posttonic syllable sets are merged with each other at $\tau = 0.6$,
and with the initial stressed and medial stressed syllables at $\tau =
0.4$.  But there is no greater similarity between this cluster and the
cluster of inital pretonic and plain syllables (formed at $\tau =
0.4$) than there is between it and the cluster of final and
monosyllables.  All three are merged into a single cluster at $\tau =
0.3$.

\subsubsection{Generalisation of Multi-Syllable OFS Model for English}\label{sec:English}

In the cluster tree for English (Figure~\ref{fig:english-clusters}),
there are clusters clearly correlated with stress and clusters clearly
correlated with position.  At $\tau = 0.3$ three clusters are formed,
one containing all medial syllable sets except the stressed medial
syllables, another containing all final syllable sets except the
stressed final syllables, and the third containing two stressed
syllable sets.  At $\tau = 0.25$, all stressed syllables together form
one cluster.  However, at $\tau = 0.2$, two unstressed syllable sets
are added to this cluster, while all the remaining unstressed sets
form the other large cluster.  Thus, in English, both stress and
position are strong determinants of phonotactic variation, but
differences resulting from stress are more pronounced than those
resulting from position.

\subsection{Discussion}

The {\sc lip} approach implemented with \OFS Modelling proceeds in
three steps.  First, the factors likely to produce phonotactic
idiosyncracy (stress and position in the above examples), and the
constituents to be used in the analysis (syllables only in the above
examples), are decided, and a prototype model is constructed on this
basis.  This prototype distinguishes as many objects at level~0 as
there are possible combinations of factors and lowest-level
constituents.  All ways in which these objects can combine to form
higher-level constituents are encoded at the corresponding higher
levels in the model.

In the second step, the prototype is instantiated with data sets from
different languages.  The degree to which the instantiated models
generalise over the given data is determined by the number of
constituents and subcategories of constituents distinguished in the
prototype.  As an example, consider the different degrees to which
three models that discriminate different numbers of syllable classes
generalise over given data.  All three models define words as
sequences of syllables, and syllables as sequences of phonemes.  The
first model has only one syllable class, the second distinguishes four
classes reflecting position in a word, and the third is the same as
the model presented in the preceding section, i.e.\ distinguishes
twelve syllable classes.  After instantiation with the same data set
of German phonological word forms from {\sc celex} used previously,
the three models will encode supersets of the data set that generalise
over it to different degrees.  Looking at subsets of words of the same
length gives some impression of the differences.  For instance, model
1 encodes $10,598$ monosyllabic German words (the total number of
different syllables in the data), whereas models 2 and 3 encode only
$6,841$ monosyllables (the actual number of monosyllabic words in {\sc
celex}).  The following table shows the number of bisyllabic words
each model encodes.\\

\hspace{-.35cm}\begin{tabular}{lr}
Model                                    & Bisyllabic words \\
\hline
(1) $Syll \hspace{.1cm} Syll$                           & $1.12 \times 10^8$ \\
(2) $Syll\_ini \hspace{.1cm} Syll\_fin$                 & $2.67 \times 10^7$ \\
(3) $( Syll\_ini\_pr \hspace{.1cm} Syll\_fin\_st ) +$               & \\
\hspace{.75cm}$( Syll\_ini\_st \hspace{.1cm} Syll\_fin\_po )$                 & $1.89 \times 10^7$ \\
\hline
$Attested \hspace{.1cm} forms$        & $7.09 \times 10^4$ \\
\end{tabular}\\

Model~3 permits about 266 times as many bisyllabic word forms as there
are in {\sc celex}, model~2 encodes 1.4 times as many as model~3, and
model~1 encodes 4.2 times as many as model~2.  Thus, through
progressively finer grained subcategories of syllables, progressively
closer approximations of the set of attested forms can be achieved.

However, doing this in an indiscriminate, language-independent way may
produce some syllable classes that are very similar.  With {\sc
os}-generalisation, the most similar classes can be merged, so that
only strongly marked differences are preserved.  However, setting
$\tau$ to any specific value is problematic.  Producing cluster trees
with a range of $\tau$ values can give some idea of important class
distinctions, and can be used as a basis for determining an
appropriate $\tau$ value.  $\tau$ can further be motivated by
different linguistic assumptions and the intended purpose of the
generalised models. Generalising different instantiations of the same
prototype for the same $\tau$ value, makes it possible to compare the
relative markedness of phonotactic variation in different languages.

\section{Summary and Further Research}

This paper described how \OFS modelling and the multi-syllable
approach can be combined with language-independent prototyping to
create a method for designing phonotactic models that (i) facilitates
automatic model construction, (ii) produces models that are
arbitrarily close approximations of the set of wellformed phonological
words in a given language, and (iii) provides a generalisation method
with control over the degree to which final models fit given data.
Extensions of the approach currently under investigation include
stochastic \OFS models, and the integration of \OFS models into
finite-state syntactic grammars.

\bibliographystyle{acl}

\bibliography{thesis,2000-sub}

\end{document}